\documentclass[10pt,journal,compsoc]{IEEEtran}
\usepackage{amsmath}
\usepackage{amsfonts}
\usepackage{mathrsfs}
\usepackage{amssymb}
\usepackage{url}
\usepackage{pgfplots}
\usepackage{subfigure}
\usepackage{multirow}
\usepackage{latexsym}
\usepackage{booktabs}
\usepackage{bm}
\usepackage{soul}
\usepackage{longtable}
\usepackage{arydshln}
\usepackage[encapsulated]{CJK}

\ifCLASSOPTIONcompsoc
  \usepackage[nocompress]{cite}
\else
  \usepackage{cite}
\fi
\ifCLASSINFOpdf
\else
\fi
\begin{document}

\title{Explicit Sentence Compression for Neural Machine Translation}

\author{Zuchao Li, Rui Wang, Kehai Chen, Masao Utiyama, Eiichiro Sumita, \\
	Zhuosheng Zhang, and Hai Zhao
	\IEEEcompsocitemizethanks{
		\IEEEcompsocthanksitem Z. Li, Z. Zhang, H. Zhao are with the Department of Computer Science and Engineering, Shanghai Jiao Tong University, Shanghai,
China, 200240. R. Wang, K. Chen, M. Utiyama and E. Sumita are with National Institute of Information and Communications Technology (NICT), Kyoto, Japan.\protect\\
		E-mail: charlee@sjtu.edu.cn.
	}
}

\markboth{Journal of \LaTeX\ Class Files,~Vol.~XX, No.~X, December~2019}%
{Li \MakeLowercase{\textit{et al.}}: Explicit Sentence Compression for Neural Machine Translation}

\IEEEtitleabstractindextext{%
\begin{abstract}
State-of-the-art Transformer-based neural machine translation (NMT) systems still follow a standard encoder-decoder framework, in which source sentence representation can be well done by an encoder with self-attention mechanism. Though Transformer-based encoder may effectively capture general information in its resulting source sentence representation, the backbone information, which stands for the gist of a sentence, is not specifically focused on. In this paper, we propose an explicit sentence compression method to enhance the source sentence representation for NMT. In practice, an explicit sentence compression goal used to learn the backbone information in a sentence. We propose three ways, including backbone source-side fusion, target-side fusion, and both-side fusion, to integrate the compressed sentence into NMT. Our empirical tests on the WMT English-to-French and English-to-German translation tasks show that the proposed sentence compression method significantly improves the translation performances over strong baselines.
\end{abstract}

}

\maketitle

\IEEEdisplaynontitleabstractindextext

\IEEEpeerreviewmaketitle

\IEEEraisesectionheading{\section{Introduction}\label{sec:introduction}}

\IEEEPARstart{N}eural machine translation (NMT) is popularly implemented as an encoder-decoder framework \cite{vaswani2017attention}, in which the encoder is right in charge of source sentence representation.
Typically, the input sentence is implicitly represented as a contextualized source representation through deep learning networks.
By further feeding the decoder, the source representation is used to learn dependent time-step context vectors for predicting target translation~\cite{bojar-etal-2018-findings}. 

In state-of-the-art Transformer-based encoder, self-attention mechanisms are good at capturing the general information in a sentence \cite{dou2018exploiting,wang2019exploiting,yang2019context}. However, it is difficult to distinguish which kind of information lying deeply under the language is really salient for learning source representation. Intuitively, when a person reads a source sentence, he/she often selectively focuses on the basic sentence meaning, and re-reads the entire sentence to understand its meaning completely.
Take the English sentence in Table \ref{tab:sc_example} as an example.
We manually annotate its basic meaning as a shorter sequence of words than in the original sentence, called backbone information.
Obviously, these words with the basic meaning contain more important information for human understanding than the remaining words in the sentence.  
We argue that such backbone information is also helpful for learning source representation, and is not explicitly considered by the existing NMT system to enrich the source sentence representation.

In this paper, we propose a novel explicit sentence compression approach to enhance the source representation for NMT.
To this end, we first design three sentence compression models to accommodate the needs of various languages and scenarios, including supervised, unsupervised, and semi-supervised ways, to learn a backbone information words sequence (as shown in Table \ref{tab:sc_example}) from the source sentence. 
We then propose three translation models, including backbone source-side fusion based NMT (\textbf{BSFNMT}), backbone target-side fusion (\textbf{BTFNMT}), and both-side fusion based NMT (\textbf{BBFNMT}), to introduce this backbone knowledge into the existing Transformer NMT system for improving translation predictions.
Empirical results on the WMT14 English-to-German and English-to-French translation tasks show that the proposed approach significantly improves the translation performance over the strong even state-of-the-art NMT baselines\footnote{Our code is available at \url{https://github.com/bcmi220/esc4nmt}.}.

\begin{table*}
	\centering 
	\caption{\label{tab:sc_example} An example of sentence compression.}
	\begin{tabular}{l|p{9.7cm}}
		\hline
		\hline
		{\bf Sentence }
		& Both the {\color{red}US} {\color{red}authorities} {\color{red}and} the {\color{red}Mexican} security {\color{red}forces} are engaged in an ongoing {\color{red}battle against} the {\color{red}drug cartels}. \\ 
		\hline
		{\bf Basic Meaning} & US authorities and Mexican forces battle against drug cartels \\
		\hline
		{\bf Backbone supervised ESC} & US and Mexican fight drug cartels \\
		\hline
		{\bf Backbone unsupervised ESC} & US authorities and Mexican security forces battle drug cartels \\
		\hline
		{\bf Backbone semi-supervised ESC} & US authorities and Mexican security forces battle against drug cartels \\
		\hline
		\hline
	\end{tabular}
\end{table*}

\section{Explicit Sentence Compression}
Generally, sentence compression\footnote{There are many types of sentence compression. In this paper, we focus on abstract sentence summarization.} is a typical sequence generation task which aims to maximize the absorption and long-term retention of large amounts of data over a relatively short sequence for text understanding~\cite{Knight:2002:SBS:604203.604207,che2015sentence}. To distinguish the importance of words in the sentence and, more importantly, to dig out the most salient part in the sentence representation, we utilize the sentence compression method to explicitly distill the key knowledge that can retain the key meaning of the sentence, termed explicit sentence compression (\textbf{ESC}) in this paper. Depending on whether or not the sentence compression is trained using \textbf{human annotated data}, the proposed method can be implemented in three ways: supervised ESC, unsupervised ESC, and semi-supervised ESC.  

\subsection{Supervised ESC} 
Sentence compression usually relies on large-scale raw data together with their human-labeled data, which can be viewed as supervision, to train a sentence compression model~\cite{rush-etal-2015-neural,hu-etal-2015-lcsts,chopra-etal-2016-abstractive,cheng-lapata-2016-neural,nallapati2016abstractive,duan-etal-2019-zero}. For example, \cite{nallapati2016abstractive} proposed an attentive encoder-decoder recurrent neural network (RNN) to model abstractive text summarization. \cite{song2019mass} furture proposed MAsked Sequence to Sequence pre-training (MASS) for the encoder-decoder sentence compression framework which reported state-of-the-art performance on both the Gigaword Corpus and DUC Corpus\footnote{\url{https://duc.nist.gov/duc2004/tasks.html}}.

Sentence compression can be conducted by a typical sequence-to-sequence model. The encoder represents the input sentence $S$ as a sequence of annotation vectors, and the decoder depends on the attention mechanism to learn the context vector for generating a compressed version $S^{'}$ with the key meaning of the input sentence.  
Recently, the new Transformer architecture proposed by \cite{vaswani2017attention}, which fully relies on self-attention networks, has exhibited state-of-the-art translation performance for several language pairs. We follow this practice and attempt to apply the Transformer architecture to such a compression task. 

\subsection{Unsupervised ESC}
A major challenge in supervised sentence compression is the scarce high quality human annotated parallel data. In practice, due to the lack of parallel annotated data, the supervised sentence compression model cannot be trained or the annotated data domain is different, resulting in the sentence compression model trained on the in-domain performing poorly on the out-of-domain.

Supervised sentence compression models have achieved impressive performances based on large corpora containing pairs of verbose and compressed sentences with human annotation~\cite{nallapati2016abstractive,song2019mass}. However, the effectiveness relies heavily on the availability of large amounts of parallel original and human-annotated compressed sentences. This hinders the sentence compression approach from further improvements for many low-resource scenarios. Recently, motivated by recent progress in unsupervised cross-lingual embeddings, the unsupervised NMT~\cite{artetxe2017unsupervised,lample2017unsupervised,lample2018phrase} opened the door to solving the problem of sequence-to-sequence learning without any parallel sentence pairs. It takes advantage of the lossless (ideal situation) nature of machine translation between languages; i.e., it can translate language $L_1$ to language $L_2$ and back translate $L_2$ to language $L_1$.
However, sentence compression does not have this feature. It is lossy from sentence $S$ to sentence $S^{'}$, which makes it difficult to restore from the compressed sentence $S^{'}$ to the original sentence $S$. 

\cite{fevry2018unsupervised} added noises to extend the original sentences and trained a denoising auto-encoder to recover the original, constructing an end-to-end training network without any examples of compressed sentences in sequence to sequence framework. In doing so, the model has to exclude and reorder the noisy sentence input, and hence learns to output more semantic important, shorter but grammatically correct sentences. There are two types of noise used in the model: \textbf{Additive Sampling Noise} and \textbf{Shuffle Noise}.

\noindent \textbf{Additive Sampling Noise:} To extend the original sentence, we sample additional sentence from the training dataset randomly, and then sub-sample a subset of words from each without replacement. The newly sampled words are appended to the original sentence.

\noindent \textbf{Shuffle Noise:} In order for the model to learn to rephrase the input sentence to make the output shorter, we shuffle the resultant additive noisy sentence.

To gain a better quality for the compressed sentences, we transfer the method of \cite{fevry2018unsupervised} into the Transformer architecture instead of their suggested RNN architecture, which makes it conducive to deeper network training and a larger corpus.

\subsection{Semi-supervised ESC}

As pointed out in \cite{song2019mass}, sequence to sequence framework has attracted much attention recently due to the advances of deep learning by using large-scale data. Many language generation tasks have only a small scale of pair data which can't support to train a deep model with good generalization ability. In comparison, there is a lot of unpaired data which is earier to obtain.

We observe a performance degradation caused by different domains in the supervised ESC. According to the experimental results of \cite{fevry2018unsupervised}, the accuracy of the unsupervised ESC is currently lower than the supervised one. Therefore, we have further adopted the semi-supervised explicit sentence compression model to alleviate this problem. Specifically, the unsupervised training (often referred to as pre-training) is performed on the unpaired data first and fine-tuning with the small scale paired data (supervised training) to obtain the ESC model with good performance and generalization ability.

\subsection{Compression Rate Control}
Explicit compression rate (length) control is a common method which has been used in previous sentence compression works. \cite{kikuchi2016controlling} examined several methods of introducing target output length information, and found that they were effective without negatively impacting summarization quality. \cite{fan2018controllable} introduced a length marker token that induces the model to target an output of a desired length, coarsely divided into discrete bins. \cite{fevry2018unsupervised} augmented the decoder with an additional length countdown input which is a single scalar that ticks down to $0$ when the generation reached the desired length.

Different with the length marker or length countdown input, to induce our model to output the compression sequence with desired length, we use beam search during generation to find the sequence $S^{'}$ that maximizes a score function $s(S^{'}, S)$ given a trained ESC model. The length normalization is introduced to account for the fact that we have to compare hypotheses of different length. Without some form of length-normalization regular $ln$, beam search will favor shorter sequences over longer ones on average since a negative log-probability is added at each step, yielding lower (more negative) scores for longer sentences. Moreover, a coverage penalty $cp$ is also added to favor the sequence that cover the source sentence meaning as much as possible according to the attention weights \cite{wu2016google}.
\begin{align}
s(S^{'}, S) &= log(P(S^{'}|S)) / ln(S^{'}) + cp(S; S^{'}),\\
ln(S^{'}) &= (5 + |S^{'}|)^{\alpha} / (5 + 1)^{\alpha}, \\
cp(S; S^{'}) &= \beta \times \sum_{i=1}^{|S|} \log(\min(\sum_{j=1}^{|S^{'}|}p_{i,j}, 1.0)),
\end{align}
where $p_{i,j}$ is the attention probability of the $j$-th target word on the $i$-th source word. Parameters $\alpha$ and $\beta$ control the strength of the length normalization and the coverage penalty. Although $\alpha$ can be used to control the compression ratio softly, we use the compression ratio $\gamma$ to control the maximum length of decoding generation by hard requirements. When the decoding length $|S^{'}|$ is greater than $\gamma |S|$, the decoding stops.

\begin{figure*}[thb!]
	\subfigure{
		\begin{minipage}[b]{0.48\linewidth}
			\centering
			\includegraphics[width=2.65in, height=3.7in]{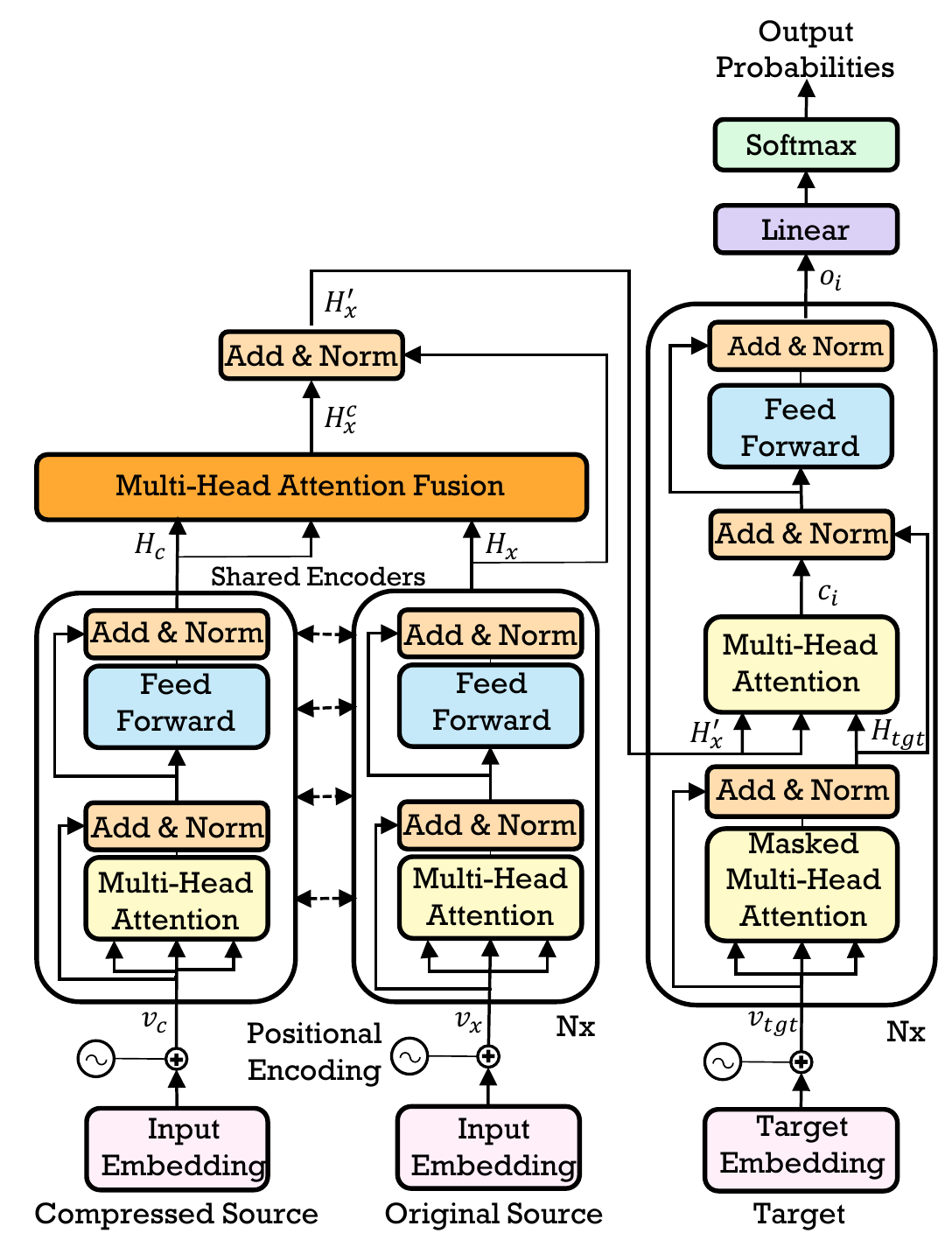}
			\caption{The architecture of proposed BSFNMT model.}
			\label{fig:bsfnmt}
		\end{minipage}
	}
	\subfigure{
		\begin{minipage}[b]{0.48\linewidth}
			\centering
			\includegraphics[width=2.65in, height=3.7in]{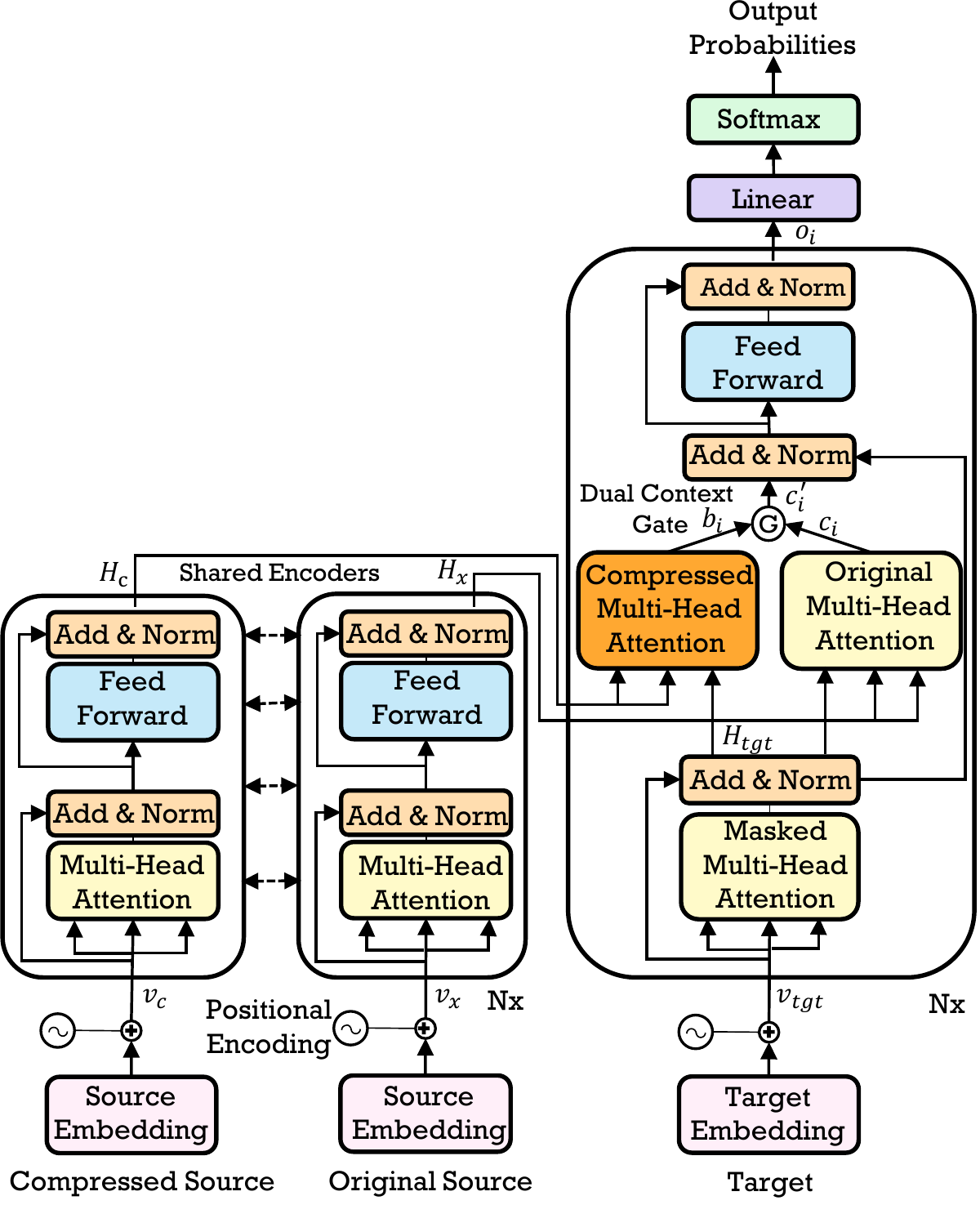}
			\caption{The architecture of proposed BTFNMT model.}
			\label{fig:btfnmt}
		\end{minipage}
	}
\end{figure*}

\section{NMT with ESC}
In this section, we first introduce the Transformer networks for machine translation.
Then based on the fusion position of the backbone knowledge sequence, we propose three novel translation models: the backbone source-side fusion based NMT model (as shown in Figure~\ref{fig:bsfnmt}), the backbone target-side based NMT model (as shown in Figure~\ref{fig:btfnmt}), and the backbone both-side based NMT.
All of these models can make use of the source backbone knowledge generated by our sentence compression models.

\subsection{Transformer Networks}
A Transformer NMT model consists of an encoder and a decoder, which fully rely on self-attention networks (\textbf{SANs}), to translate a sentence in one language into another language with equivalent meaning.
Formally, one input sentence $x$=$\{x_1, \cdots, x_J\}$ of length $J$ is first mapped into a sequence of word vectors. Then the sequence and its position embeddings add up to form the input representation $v_x=\{v^x_1, \cdots, v^x_J\}$.
The sequence $\{v^x_1, \cdots, v^x_J\}$ is then packed into a query matrix $\textbf{Q}_x$, a key matrix $\textbf{K}_x$, and a value matrix $\textbf{V}_x$.
For the SAN-based encoder, the self-attention sub-layer is first performed over $\textbf{Q}$, $\textbf{K}$, and $\textbf{V}$ to the matrix of outputs as:
\begin{equation}
\textup{SelfAtt}(\textbf{Q}, \textbf{K}, \textbf{V})=\textup{Softmax}(\frac{\textbf{Q}\textbf{K}^{T}}{\sqrt{d_{model}}})\textbf{V},\\
\label{eq1:self-attention}
\end{equation}
where $d_{model}$ represents the dimensions of the model. Similarly, the translated target words are used to generate the decoder hidden state $\textbf{s}_i$ at the current time-step $i$.
Generally, the self-attention function is further refined as multi-head self-attention to jointly consider information from different representation subspaces at different positions: 
\begin{equation}
\begin{split}
&  \textup{MultiHead}(\textbf{Q}, \textbf{K}, \textbf{V}) =\textup{Concat}(\textup{head}_{1}, \cdots, \textup{head}_H)\textbf{\textit{W}}^{O},\\
& \;\;\;\;\;\;\;\;\;\;\;\;\;\;\;\;\;\;\;\;\;\textup{head}_h=\textup{SelfAtt}(\textbf{Q}\textbf{W}_{h}^{Q},\textbf{k}\textbf{W}_{h}^{K},\textbf{V}\textbf{W}_{h}^{V}), 
\end{split}
\label{eq2:multiHead_self-attention}
\end{equation}
where the projections are parameter matrices $\textbf{W}_{h}^{Q}$$\in$$\mathbb{R}^{d_{model}\times d_k}$,  $\textbf{W}_{h}^{K}$$\in$$\mathbb{R}^{d_{model}\times d_k}$, $\textbf{W}_{h}^{V}$$\in$$ \mathbb{R}^{d_{model}\times d_v}$, and $\textbf{W}^{O}$$\in$$\mathbb{R}^{hd_{v}\times d_{model}}$.
For example, there are $H$=8 heads, $d_{model}$ is 512, and $d_k$=$d_v$=512/8=64.
A position-wise feed-forward network (FFN) layer is applied over the output of multi-head self-attention, and then is added with the matrix $\textbf{V}$ to generate the final source representation $H_{x}$=$\{H^{x}_1, \cdots, H^{x}_J\}$:
\begin{equation}
H_{x} = \textup{FFN}(\textup{MultiHead}(\textbf{Q}, \textbf{K}, \textbf{V}))+\textbf{V}.
\label{eq3:PositionWiseFeedForward}
\end{equation}

The SAN of decoder then uses both $H_x$ and target context hidden state $H_{tgt}$ to learn the context vector $o_i$ by ``encoder-decoder attention":
\begin{equation}
c_i = \textup{FFN}(\textup{MultiHead}(H_{tgt}, H_x, H_x)),
\label{eq4:context_self-attention}
\end{equation}
\begin{equation}
o_i = c_i + H_{tgt}.
\label{eq:tgt_fusion}
\end{equation}

Finally, the context vector $o_{i}$ is used to compute translation probabilities of the next target word $\textit{y}_i$ by a linear, potentially multi-layered function:
\begin{equation}
P(\textit{y}_i | \textit{y}_{<i}, \textit{x}) \propto 
\textup{Softmax}(\textbf{L}_\textit{o} \textbf{GeLU}(\textbf{L}_\textit{w}o_{\textit{i}})),
\label{eq5:Probabilities}
\end{equation}
where $\textbf{L}_{o}$ and $\textbf{L}_{w}$ are projection matrices.

\subsection{Backbone Source-side Fusion based NMT}
In the backbone source-side fusion based NMT (\textbf{BSFNMT}) model, given an input sentence $x$=$\{x_1, \cdots, x_J\}$, there is an additional compressed sequence $x_c$=$\{x^c_1, \cdots, x^c_K\}$ of length $K$ generated by the proposed sentence compression model.
This compressed sequence is also input to the SAN shared with the original encoder with word vectors $v_c = \{v^c_i, \cdots, v^c_K\}$ in shared vocabulary to learn its final representation $H_{c}$=$\{H^{c}_1, \cdots, H^{c}_K\}$.
In the proposed SFNMT model, we introduce an additional multi-head attention layer to fuse the compressed sentence and the original input sentence for learning a more effective source representation.

Specifically, for the multi-head attention-fusion layer, a compressed sentence-specific context representation $H_x^c$ is computed by the multi-head attention on the original sentence representation $H_x$ and the compressed sentence representation $H_c$:
\begin{equation}
\begin{split}
H_x^c = \textup{FFN}(\textup{MultiHead}(H_x, H_c, H_c)).
\end{split}
\end{equation}

$H_x^c$ and $H_x$ are added to form a fusion source representation $H_{x}^{'}$: 
\begin{equation}
H_{x}^{'} = H_x +H_x^c.
\label{Eq9:fuse_src_rep}
\end{equation}

Finally, the $H_{x}^{'}$ instead of $H_{x}$ is input to the Eq.~\eqref{eq4:context_self-attention} in turn for predicting the target translations word by word.

\subsection{Backbone Target-side Fusion based NMT}
In the backbone target-side fusion based NMT (\textbf{BTFNMT}) model, both the original sentence and its compressed version are also represented as $H_x$ and $H_c$ respectively by the shared SANs.
We then use a tuple ($H_x,H_c$) instead of the source-side fusion representation $H_x^{'}$ as the input to the decoder.
Specifically, we introduce an additional ``encoder-decoder attention" module into the decoder to learn the compressed sequence context $b_i$ at the current time-step $i$:
\begin{equation}
b_i=\textup{FFN}(\textup{MultiHead}(H_{tgt}, H_c, H_c)).
\label{Eq11:skeleton_ctx}
\end{equation}

Since we are here to treat the original sentence and the compressed sentence as two independent source contexts when encoding at the source side, we use a context gate $g_c$ for integrating two independent contexts of the source: original context $c_i$ and compressed context $b_i$. The gate $g_i$ is calculated by:
\begin{equation}
g_i = \sigma(\textbf{MLP}([c_i; b_i])).
\end{equation}

Therefore, the final target fusion context $c_i'$ is:
\begin{equation}
c_i' = g_i \otimes c_i + (1 - g_i) \otimes b_i,
\end{equation}
where $\sigma$ is the logistic sigmoid function, $\otimes$ is the point-wise multiplication, and $[\cdot]$ represent the concatenation operation.

The context $c_i'$ is input to replace the $c_i$ the Eq.~\eqref{eq:tgt_fusion} to compute the probabilities of next target word.

\subsection{Backbone Both-side Fusion based NMT}
In the backbone both-side fusion based NMT (\textbf{BBFNMT}) model, we combine \textbf{BSFNMT} and \textbf{BTFNMT}. Both the original representation $H_x$ and its compressed enhanced representation $H_x^{'}$ are as the input to the decoder. Similarly, we introduce an additional ``encoder-decoder attention" module into the decoder to learn the compressed sequence enhanced context $b_i^{'}$ at the current time-step $i$:
\begin{equation}
b_i^{'}=\textup{FFN}(\textup{MultiHead}(H_{tgt}, H_{x}^{'}, H_{x}^{'})).
\end{equation}

Then, the context gate $g_i$ consistent with \textbf{BTFNMT} is applied to combine the two context information $c_i$ and $b_i^{'}$.

\section{Experiments}
\subsection{Setup}
\paragraph{Sentence Compression} 
To evaluate the quality of our sentence compression model, we used the Annotated Gigaword corpus~\cite{napoles2012annotated} as the benchmark~\cite{rush2015neural}. 
The data includes approximately 3.8 M training samples, 400 K validation samples, and 2 K test samples. The byte pair encoding (BPE) algorithm~\cite{sennrich2016neural} was adopted for subword segmentation, and the vocabulary size was set at 40 K for our supervised, unsupervised and semi-supervised settings \cite{ZhangEffective}. 

Baseline systems include AllText and F8W \cite{rush2015neural,wang2018learning}. 
F8W is simply the first 8 words of the input, and AllText uses the whole text as the compression output. The $F_1$ score of ROUGE-1 (R-1), ROUGE-2 (R-2), and ROUGE-L (R-L) was used to evaluate this task~\cite{lin2004rouge}. We use beam search with a beam size of 5, the length length normalization of 0.5, and the coverage penalty of 0.2. 

For the semi-supervised setting, in order to make the results comparable to ~\cite{song2019mass}, we used the same 190M English monolingual unpaired data from WMT News Crawl datasets for pre-training (unsupervised training). We included the other pretraining methods: masked language modeling (MLM, BERT) \cite{devlin2018bert}, denoising auto-encoder (DAE) \cite{vincent2008extracting}, and masked sequence to sequence (MASS) \cite{song2019mass} to compare with our unsupervised pretraining method in the semi-supervised setting.

\paragraph{Machine Translation} 
The proposed NMT model was evaluated on the WMT14 English-to-German (EN-DE) and English-to-French (EN-FR) tasks, which are both standard large-scale corpora for NMT evaluation.
For the EN-DE translation task, 4.43 M bilingual sentence pairs from the WMT14 dataset were used as training data, including Common Crawl, News Commentary, and Europarl v7. 
The newstest2013 and newstest2014 datasets were used as the dev set and test set, respectively.
For the EN-FR translation task, 36 M bilingual sentence pairs from the WMT14 dataset were were used as training data.
Newstest12 and newstest13 were combined for validation and the newstest14 was the test set, following the setting of \cite{gehring2017convolutional}. 
The BPE algorithm~\cite{sennrich2016neural} was also adopted, and the joint vocabulary size was set at 40 K.
For the hyper-parameters of our Transformer (base/large) models, we followed the settings used in \cite{vaswani2017attention}'s work.

In addition, we also reported the state-of-the-art results in recent literatures, including modelling local dependencies (\textbf{Localness})~\cite{yang-etal-2018-modeling}, fusing multiple-layer representations in SANs (\textbf{Context-Aware})~\cite{DBLP:journals/corr/abs-1902-05766}, and fusing all global context representations in SANs (\textbf{global-deep context})~\cite{dou-etal-2018-exploiting}.
MultiBLEU was used to evaluate the translation task. 

\subsection{Main Results}
\paragraph{Sentence Compression} 

\begin{table}[h]
	\centering
	\caption{\label{tab:sc_results}Performance on the sentence compression task}
	\begin{tabular}{l c c c}
		\hline
		\hline
		\textbf{Model}& \textbf{R-1}& \textbf{R-2} & \textbf{R-L} \\
		\hline
		\textit{Baselines:} & & & \\
		All text & 28.91 & 10.22 & 25.08\\
		F8W & 26.90 & 9.65 & 25.19 \\
		\hline
		\textit{Unsupervised:} & & & \\
		~\cite{fevry2018unsupervised} & 28.42 & 7.82 & 24.95 \\
		\textbf{ESC (This work)} & 31.37 & 8.25 & 28.01 \\
		\hline
		\textit{Supervised:} & & & \\
		RNN-based Seq2seq & 35.50 & 15.54 & 32.45 \\
		\cite{nallapati2016abstractive} & 34.97 & 17.17 & 32.70 \\
		\textbf{ESC (This work)} & 37.53 & 18.48 & 34.79 \\
		\hline
		\textit{Semi-supervised:} & & & \\
		MLM Pretraining & 37.75 & 18.45 & 34.85 \\
		DAE Pretraining & 35.97 & 17.17 & 33.14 \\
		\cite{song2019mass} & 38.73 & 19.71 & 35.96 \\
		\textbf{ESC (This work)} & 39.54 & 20.35 & 36.79 \\
		\hline
		\hline
	\end{tabular}
\end{table}

To evaluate the quality of our sentence compression model, we conducted a horizontal comparison between the proposed sentence compression model and other sentence compression models in different settings.
Table~\ref{tab:sc_results} shows the comparison results. 
We observed that the proposed unsupervised ESC model performed substantially better than Fevry and \cite{fevry2018unsupervised}'s unsupervised method.
The proposed supervised ESC model also substantially outperformed the RNN-based Seq2seq and \cite{nallapati2016abstractive}'s baseline method.
That is, our supervised model gave +2.0 improvements on R-1, R-2, and R-L scores over the RNN-based Seq2seq.
This means that the proposed Transformer-based approaches can generate compressed sentences of high quality. 

We further compared our semi-supervised model with the semi-supervised pretraining methods of MLM \cite{devlin2018bert}, DAE \cite{vincent2008extracting}, and MASS \cite{song2019mass}. Our unsupervised pretrainining method outperformed the other unsupervised pretrainining ones on the sentence compression task consistently.

\begin{table*}[h]
	\centering
	\caption{\label{tab:mt_main_results}Comparison with existing NMT systems on WMT14 EN-DE and EN-FR Translation Tasks. ``++/+" after the BLEU score indicate that the proposed method was significantly better than the corresponding baseline Transformer (base or big) at significance level p$<$0.01/0.05. 	``\#Speed" denotes the decoding speed measured in target tokens per second.}
	\begin{tabular}{l||l|rr||l|rr}
			\hline
			\hline
			\multicolumn{1}{c||}{\multirow{1}{*}{System}} &  \multicolumn{1}{c}{\multirow{1}{*}{EN-DE}} & \#Speed & \#Params & \multicolumn{1}{c}{\multirow{1}{*}{EN-FR}} & \#Speed & \#Params   \\ \hline
			\multicolumn{7}{c}{\textit{Existing NMT systems}}   \\ \hline
			Transformer (base)~\cite{vaswani2017attention}     & 27.3    & N/A  & 65.0M    & 38.1    & N/A & N/A \\
			\;\;\;+Localness~\cite{yang-etal-2018-modeling} & 28.11 & N/A & 88.8M & N/A & N/A & N/A \\
			\;\;\;+Context-Aware SANs~\cite{DBLP:journals/corr/abs-1902-05766} & 28.26 & N/A & 194.9M & N/A & N/A & N/A \\
			\;\;\;+global-deep context~\cite{dou-etal-2018-exploiting} & 28.58 & N/A & 111M & N/A & N/A & N/A \\ \hdashline
			Transformer (big)~\cite{vaswani2017attention} & 28.4  & N/A & 213.0M    & 41.0   & N/A & N/A \\ 
			\;\;\;+Localness~\cite{yang-etal-2018-modeling} & 28.89 & N/A & 267.4M & N/A & N/A & N/A \\
			\;\;\;+Context-Aware SANs~\cite{DBLP:journals/corr/abs-1902-05766} & 28.89 & N/A & 339.6M & N/A & N/A & N/A \\
			\;\;\;+global-deep context~\cite{dou-etal-2018-exploiting} & 29.21 & N/A & 396M & N/A & N/A & N/A \\
			\hline
			\multicolumn{7}{c}{\textit{Our NMT systems}}  \\ \hline
			Transformer (base)        &  27.24    & 131k & 66.5M & 38.21    & 130k & 85.7M  \\ 
			\textbf{BSFNMT}     &  27.75++    & 121k  & 72.1M  & 39.09++    & 120k & 89.0M \\ 
			\textbf{BTFNMT}     &  28.14+    & 120k & 72.7M & 39.22++    & 119k & 89.8M \\ 
			\textbf{BBFNMT} & 28.35++ & 119k & 78.6M & 39.40++ & 116k & 91.4M \\
			\hdashline
			Transformer (big)         &  28.23   & 11k & 221.0M  & 41.15    & 11k & 222.3M \\
			\textbf{BSFNMT}        &  28.52+   & 10k & 225.2M & 41.92+    & 9k & 227.1M  \\
			\textbf{BTFNMT}        &  29.16++  & 9k & 225.7M  & 42.22++    & 8k & 227.5M  \\
			\textbf{BBFNMT} & 29.37++ & 8k & 228.9M & 42.52++ & 8k & 230.3M \\
			\hline \hline
	\end{tabular}
\end{table*} 

\paragraph{Machine Translation} 
According to the results in Table~\ref{tab:sc_results}, we chose the \textbf{semi-supervised ESC model} (which performed the best) to generate compressed sentences for the machine translation task. 
The main results on the WMT14 EN-DE and EN-FR translation tasks are shown in Table~\ref{tab:mt_main_results}. 
In the EN-DE task, we made the following observations:

1) The baseline Transformer (base) in this work achieved a performance comparable to the original Transformer (base)~\cite{vaswani2017attention}.
This indicates that it is a strong baseline NMT system.

2) All \textbf{BSFNMT}, \textbf{BTFNMT}, and \textbf{BBFNMT} significantly outperformed the baseline Transformer (base/big) and only introduces a very small amount of extra parameters. 
This indicates that the learned compressed backbone information was beneficial for the Transformer translation system.

3) Among the proposed three methods, \textbf{BTFNMT} performed better than \textbf{BSFNMT}. This indicates that the backbone fusion at the target-side is better than at the source-side.
In addition, \textbf{BBFNMT} (base/big)  outperformed the comparison systems +Localness and +Context-Aware SANs.
This indicates that the compression knowledge as an additional context can enhance NMT better.

4) \textbf{BBFNMT} (based) is comparable to the +global-deep context, the best comparison system, while \textbf{BBFNMT} (big) slightly outperformed +global-deep context by $0.16$ BLEU scores.
In particular, the parameters of \textbf{BBFNMT} (base/big) model, which just increased $12.1/7.9$M over the Transformer (base/big), were only 70\% of the +global-deep context model.   
This denotes that the \textbf{BBFNMT} model is more efficient than the +global-deep context model. In addition, the training speed of the proposed models slightly decreased ($8\%$), compared to the corresponding baselines.

5) The proposed \textbf{BBFNMT} (base) slightly outperformed the Transformer (big) which contains much more parameters than \textbf{BBFNMT} (base). This indicates that our improvement is not likely to be due to the increased number of parameters. 

For the EN-FR translation task, the proposed models gave similar improvements over the baseline systems and comparing methods (except that the Transformer (big) performed much more better than  Transformer (base)). These results show that our method is robust for improving the translation of other language pairs.

\subsection{Ablation Study}
\paragraph{Evaluating Sentence Compression}

To demonstrate the effectiveness of sentence compression, we compared the compressed sentences ($\gamma = 0.6$) generated in the Transformer translation system (BBFNMT) under different settings: AllText, F8W, RandSample (random sampling), supervised ESC, Unsupervised ESC and semi-supervised ESC. 
Table~\ref{tab:KC_performance} shows the results on newstest2014 for the EN-DE translation task. 

\begin{table}[h]
	\centering
	\caption{The effect of our ESC methods. \label{tab:KC_performance}}
	\begin{tabular}{l|c}
			\hline
			\hline
			Model & BLEU on EN-DE \\ \hline
			Baseline            &   27.24   \\ 
			\;\;\;+\textbf{AllText}  & 27.24 \\
			\;\;\;+\textbf{F8W}  & 27.40 \\
			\;\;\;+\textbf{RandSample} & 26.53 \\
			\;\;\;+\textbf{Supervised ESC}   &   27.80  \\
			\;\;\;+\textbf{Unsupervised ESC} &   27.97 \\ 
			\;\;\;+\textbf{Semi-supervised ESC}   &   28.35  \\
			\hline
			\hline
	\end{tabular}
\end{table}  

We made the following observations: 1) Simply introducing AllText and F8W achieved few improvement, and RandSample is lower than the baseline. In comparison, all the  +supervised ESC, +unsupervised ESC, and +semi-supervised ESC models substantially improved the performance over the baseline Transformer (base).
This means that our ESC method provides a richer source information for machine translation tasks.

2) +Unsupervised ESC can gain better improvements over the +supervised ESC although supervised ESC model can achieve higher quality than the unsupervised ESC model in the benchmark test dataset.  This may be due to that the annotated sentence compression training data is in different domain with the WMT EN-DE traing data. Meanwhile, +Semi-supervised ESC with annotated data fine-tuning outperformed both +Unsupervised and +supervised ESC.

\paragraph{Effect of Encoder Parameters}

In our model, representations of the original sentence and its compressed version were learned by a shared encoder.
To explore the effect of the encoder parameters, we also designed a \textbf{BBFNMT} with two independent encoders to learn representations of the original sentence and its compressed version, respectively. 
Table~\ref{tab:encoder_papameter} shows results on the newstest2014 test set for the WMT14 EN-DE translation task.

\begin{table}[h]
	\centering
	\caption{The effect of encoder parameters. \label{tab:encoder_papameter}}
	\begin{tabular}{l|cc}
		\hline
		\hline
		Model & BLEU & \#Params\\
		\hline
		Transformer (base) & 27.24  & 66.4M \\ \hdashline
		\textbf{BBFNMT w/ Shared encoder} & 28.35 & 78.6M\\  
		\textbf{BBFNMT w/ Independent encoders}  &  28.50 & 91.6M\\
		\hline
		\hline
	\end{tabular}
\end{table} 

The \textbf{BBFNMT} (w/ independent params) slightly outperformed the proposed shared encoder model by a BLEU score of 0.15, but its parameters increased by approximately 30\%.
In contrast, the parameters in our model are comparable to the baseline Transformer (base).
Considering the parameter scale, we took a shared encoder to learn source representation, which makes it easy to verify the effectiveness of the additional translation knowledge, such as our backbone knowledge.

\paragraph{Evaluating Compression Ratio}

In order to verify the impact of different compression ratios on translation quality, we conducted experiments on EN-DE translation task with semi-supervised sentence compression in \textbf{BBFNMT} model.

\begin{figure}[thb!]
	\setlength{\abovecaptionskip}{0pt}
	\begin{center}
		\pgfplotsset{height=5.0cm,width=4.0cm,compat=1.14,every axis/.append style={thick}}
		\begin{tikzpicture}
		\begin{axis}
		[width=8.0cm, enlargelimits=0.13, tick align=outside, legend style={cells={anchor=west},legend pos=south east, legend row=3, columns=1,every axis legend/.append style={
				at={(1,0)}}}, xticklabels={ $0$, $0.1$,$0.2$, $0.3$, $0.4$, $0.5$, $0.6$, $0.7$, $0.8$, $0.9$, $1.0$}, 
		xtick={0,1,2,3,4,5,6,7,8,9,10}, 
		ylabel={BLEU score},xlabel={Compression Ratio},font=\small]
		
		\addplot+ [sharp plot, mark=*,mark size=1.2pt,mark options={solid,mark color=orange}, color=orange] coordinates
		{(0,27.24)(1,27.30)(2,27.41)(3,27.65)(4,27.98)(5,28.31)(6,28.35)(7,28.24)(8,28.01)(9,27.77)(10,27.62)};
		\end{axis}
		\end{tikzpicture}
		\caption{\label{fig:gamma} Performances on EN-DE newstest2014 with different sentence compression ratios.}
	\end{center}
\end{figure}
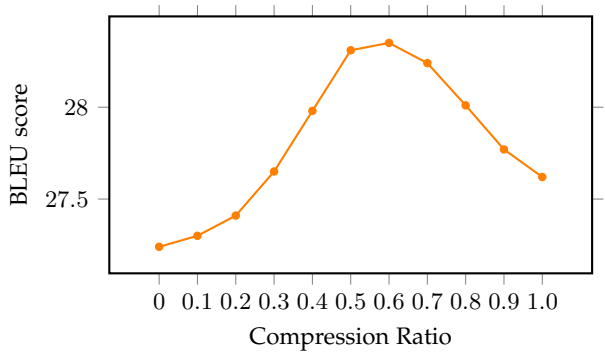

We controled the compression ratio $\gamma$ from 0 to 1.0. Consider two boundary conditions, when the compression ratio $\gamma = 0$, it means no compression sequence generated, which is the same as the vanilla Transformer. When the compression ratio $\gamma = 1.0$, it is equivalent to re-paraphrasing the source sentence using the sentence compression model (maintaining the same length) as the additional input for \textbf{BBFNMT}.

The experimental results are shown in Fig. \ref{fig:gamma}. As can be seen from the results, in our experiments, sentence compression (re-paraphrasing) can bring  performance improvement, even when the compression ratio $\gamma=1.0$ and the sentence length is not shortened, re-paraphrasing can still bring slight improvement of translation quality. On the wmt14 EN-DE translation task, the compression ratio $\gamma$ was set to 0.6 to get the best results.

\section{Related Work}

To let the translation have more focus over the source sentence information, efforts have been initiated on exploiting sentence segmentation,  sentence simplification, and sentence compression for machine translation. 
\cite{xiao2014hybrid} presented a approach to integrating the sentence skeleton information into a phrase-based statistic machine translation system. \cite{xiao2016syntactic} proposed an approach to modeling syntactically-motivated skeletal structure of source sentence for statistic machine translation.
\cite{mellebeek2006syntactic} describe an early approach to skeleton-based translation, which decomposes input sentences into syntactically meaningful chunks. The central part of the sentence is identified and remains unaltered while other parts of the sentence are simplified. This process produces a set of partial, potentially overlapping translations which are recombined to form the final translation. 
\cite{sudoh2010divide} describe a “divide and translate” approach to dealing with complex input sentences. They parse the input sentences, replace subclauses with placeholders and later substitute them with separately translated clauses. Their method requires training translation models on clause-level aligned parallel data with placeholders in order for the translation model to deal with the placeholders correctly.
\cite{pouget2014overcoming} experimented with automatically segmenting the source sentence to overcome problems with overly long sentences. 
\cite{hasler2017source} showed that the spaces of original and simplified translations can be effectively combined using translation lattices and compare two decoding approaches to process both inputs at different levels of integration. 

Different from these work, our proposed sentence compression model does not rely on any known linguistics motivated (such as syntax) skeleton simplification, but directly trains a computation motivated sentence compression model to learn to compress sentences and re-paraphrase them directly in seq2seq model. Though with a pure computation source, our sentence compression model can surprisingly generate more grammatically correct and refined sentences, and the words in the compressed sentence do not have to be the same as the original sentence. In the meantime, our sentence compression model can stably give source backbone representation exempt from unstable performance of a syntactic parser which is essential for syntactic skeleton simplification. Our sentence compression model can perform unsupervised training on large-scale data sets, and then use the supervised data for finetune, which is more promising from the results.

\section{Conclusion and Future work} 
To give a more focused source representation,  this paper makes the first attempt to propose an explicit sentence compression method to enhance state-of-the-art Transformer-based NMT.  
To demonstrate that the proposed sentence compression enhancement is indeed helpful for the neural machine translation,
We evaluate the impact of the proposed model on the large-scale WMT14 English-to-German and English-to-French translation tasks. 
The experimental results on WMT14 EN-DE and EN-FR translation tasks show that our proposed NMT model can yield significantly improved results over strong baseline translation systems. 
In the future work, we will release a pre-trained language model that uses unsupervised sentence compression as the pre-training objective to demonstrate the performance of unsupervised sentence compression in representation learning.



%



\ifCLASSOPTIONcompsoc
  \section*{Acknowledgments}
\else
  \section*{Acknowledgment}
\fi
The corresponding authors are Rui wang and Hai Zhao. Zuchao Li and Zhuosheng Zhang were internship research fellows at NICT when conducting this work.
Hai Zhao was partially supported by National Key Research and Development Program of China (No. 2017YFB0304100) and Key Projects of National Natural Science Foundation of China (No. U1836222 and No. 61733011). Rui Wang was partially supported by JSPS grantin-aid for early-career scientists (19K20354): “Unsupervised Neural Machine Translation in Universal Scenarios” and NICT tenure-track researcher startup fund “Toward Intelligent Machine Translation”.

\ifCLASSOPTIONcaptionsoff
  \newpage
\fi

\bibliographystyle{IEEEtran}
\bibliography{references}

\end{document}